\documentclass[runningheads]{llncs}
\usepackage[T1]{fontenc}
\usepackage{algorithm} 
\usepackage{algpseudocode}
\usepackage{amsmath}
\usepackage{graphicx}
\usepackage[compatibility=false]{caption}
\usepackage{subcaption}
\usepackage{float}
\usepackage{xcolor}
\usepackage{booktabs}
\usepackage{tabularx}
\usepackage{array}
\usepackage{hyperref}
\begin{document}
\titlerunning{FairEquityFL}
\title{FairEquityFL -- A Fair and Equitable Client Selection in Federated Learning for Heterogeneous IoV Networks}

\author{Fahmida Islam\thanks{The corresponding author sincerely acknowledges the generous support of the International Macquarie University Research Excellence Scholarship (Allocation Number: 20236076).}\and
Adnan Mahmood\and
Noorain Mukhtiar\and
Kasun Eranda Wijethilake\and
Quan Z. Sheng}

\authorrunning{F. Islam et al.}

\institute{School of Computing, Macquarie University, Sydney, NSW 2109, Australia 
\email{fahmida.islam1@hdr.mq.edu.au}, 
\email{adnan.mahmood@mq.edu.au},
\email{noorain.mukhtiar@hdr.mq.edu.au},
\email{kasuneranda.wijethilake@hdr.mq.edu.au},
\email{michael.sheng@mq.edu.au}} 

\maketitle%
\begin{abstract}

Federated Learning (FL) has been extensively employed for a number of applications in machine learning, i.e., primarily owing to its privacy preserving nature and efficiency in mitigating the communication overhead. Internet of Vehicles (IoV) is one of the promising applications, wherein FL can be utilized to train a model more efficiently. Since only a subset of the clients can participate in each FL training round, challenges arise pertinent to fairness in the client selection process. Over the years, a number of researchers from both academia and industry have proposed numerous FL frameworks. However, to the best of our knowledge, none of them have employed fairness for FL-based client selection in a dynamic and heterogeneous IoV environment. Accordingly, in this paper, we envisage a FairEquityFL framework to ensure an equitable opportunity for all the clients to participate in the FL training process. In particular, we have introduced a \textit{sampling equalizer module} within the \textit{selector} component for ensuring fairness in terms of fair collaboration opportunity for all the clients in the client selection process. The \textit{selector} is additionally responsible for both monitoring and controlling the clients' participation in each FL training round. Moreover, an outlier detection mechanism is enforced for identifying malicious clients based on the model performance in terms of considerable fluctuation in either accuracy or loss minimization. The \textit{selector} flags suspicious clients and temporarily suspend such clients from participating in the FL training process. We further evaluate the performance of FairEquityFL on a publicly available dataset, FEMNIST. Our simulation results depict that FairEquityFL outperforms baseline models to a considerable extent. 

\end{abstract}

\section{Introduction}

Over the past decade or so, the rapid proliferation of the Internet of Things (IoT) and vehicular ad hoc networks have transformed vehicles into \textit{smart} objects \cite{yarradoddi2022federated}. As a result, an unprecedented amount of data is being generated. This substantial amount of data can be intelligently utilized for deriving both real-time and fact-based information, and that could be indispensable for numerous vehicular safety applications, i.e., automatic emergency breaking, pedestrian detection, and the collision avoidance. Since Machine Learning (ML) can be utilized to analyze large volume of data, i.e., primarily for identifying patterns and obtaining real-time insights, utilization of ML for data consolidation and aggregation is imperative \cite{sharifani2023machine}. However, owing to the limitations pertinent to computational power, network constraints, and, in particular, the sensitivity of the data, it is not feasible by any means to transmit the raw data to the server for further processing. Therefore, it is highly imperative to process the data in a secure environment as to guarantee an individual's privacy. 

\begin{figure}[ht]
    \centering
    \includegraphics[width=1.1\textwidth]{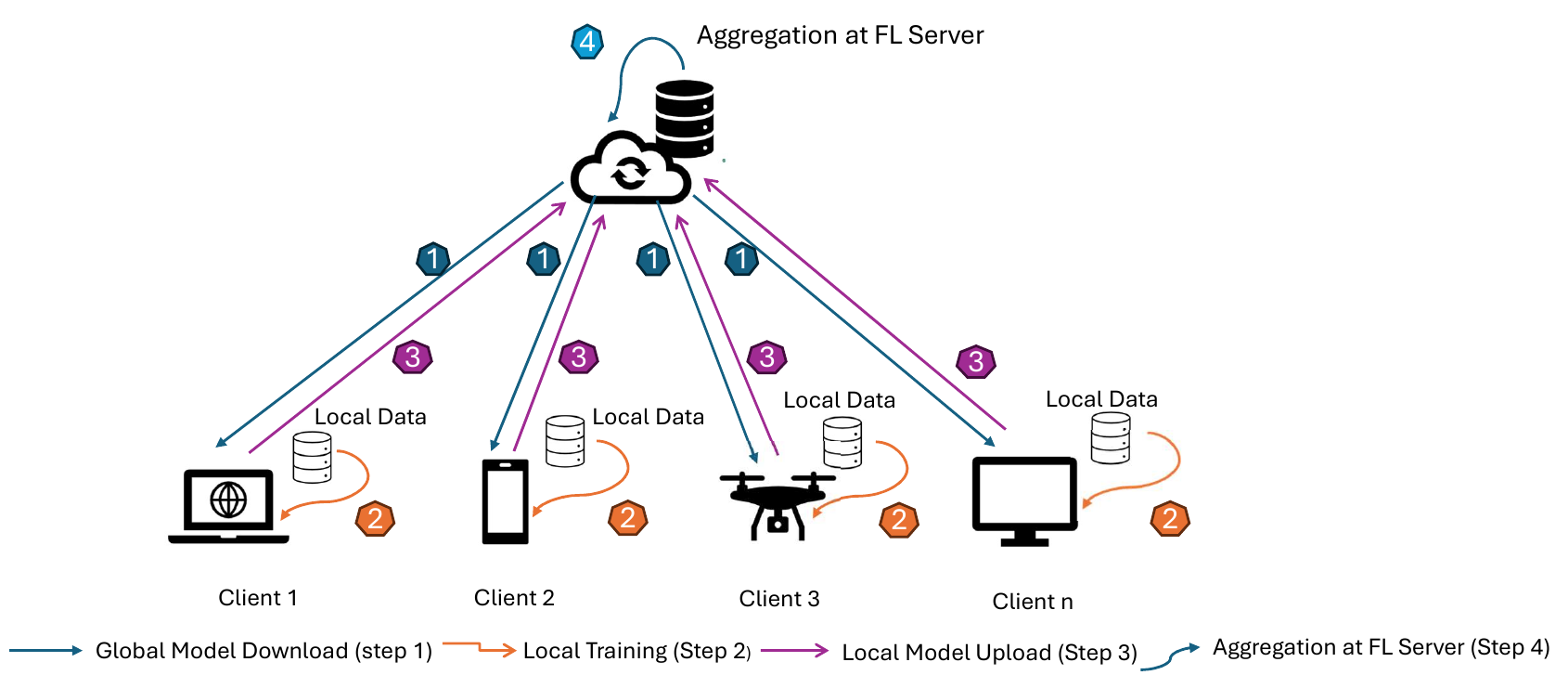}
    \caption{A classic federated learning framework.}
    \label{fig:Figure 1}
\end{figure}

Federated Learning (FL) was envisaged by Google in May 2016 as a privacy preserving framework, wherein data is locally stored at the client, multiple clients collaborate by sharing their local gradients, and global server aggregates the local gradients to train the global model \cite{mcmahan2017communication}. \hyperref[fig:Figure 1]{Figure~1} depicts the key steps in a classic FL approach. At step 1, the FL server initiates the process by sending the global model to the participating clients, i.e., the ones which are thus randomly selected. At step 2, the said clients train their respective local models by utilizing the local data. At step 3, the clients send their trained model parameters' updates to the server. Finally, at step 4, the FL server aggregates the trained local parameters received from all the participating clients and develops a new global model. This process continues until the FL model achieves desired accuracy or until reaching to predetermined rounds. 

In the realm of the Internet of Vehicles (IoV), FL can contribute to a number of scenarios, i.e., autonomous driving, planning of the road safety measures, and processing of the traffic images \cite{du2020federated}\cite{zhao2022federated}. Nevertheless, besides several advantages, it is posed with numerous challenges owing to the heterogeneity among the participants in terms of their respective storage, computing power, network strength, communication capability, and robustness. 

It is also pertinent to note that owing to several constraints, not all the clients are able to participate in the FL model training in a concurrent round. Instead, only a subset of the clients are able to take part within a single communication round \cite{fu2023client}. Therefore, FL models tend to reflect a biased result if all of the clients do not get an opportunity to participate in the FL model training since some of the clients may participate quite frequently, whereas, others may get overlooked for a longer period of time \cite{gouissem2023comprehensive}. Hence, it is indispensable to develop a client selection strategy which guarantees an opportunity for all the clients to participate in the FL model training without compromising a model's performance in terms of the accuracy level, loss minimization, and time-to-convergence \cite{zhu2022online}. Accordingly, to achieve fairness in the FL client selection process, it is crucial to restrict frequent participation of the same clients and to not overlook a certain client for a longer period of time. Moreover, malicious clients may take part in the FL model training and share the malicious local model updates to the server. Subsequently, the global model would be impacted and it would, in turn, negatively affect the local model training of the other clients \cite{cao2022mpaf}, thereby resulting in the data security and integrity issues. Therefore, it is imperative to have an intelligent mechanism for timely mitigating the risks of malicious clients' participation. 

As only a portion of the clients take part during each communication round, it is crucial that the clients are selected in a manner which reflects a fair collaboration of all the clients \cite{shi2023fairness}. If the server prioritizes the clients offering a shorter response time or possessing a larger dataset, some of the clients would be skipped and the global model would suffer a performance degradation \cite{zhu2022online}. However, there are numerous aspects that should be taken into consideration for the selection of the clients within each communication round, e.g., computational resource and client availability. Hence, it becomes challenging to incorporate fairness into the client selection process. Different researchers have suggested to consider different aspects for the client selection process. It is, nevertheless, of the essence that all of the clients participate in the FL model training so as to ensure that the model can reflect the majority of the clients' contribution. Moreover, if all of the clients get an opportunity to participate, their would be a higher chance of achieving more accurate and reliable results. Furthermore, ensuring participation opportunity for all the clients in the FL model training guarantees that equity prevails and that malicious clients would fail to produce a biased result since they would not be participating too often. 

Over the past decade or so, a considerable amount of research has been conducted for incorporating fairness into the client selection process \cite{shi2023towards}. However, to the best of our knoweldge, none of them have employed fairness for FL-based client selection in a dynamic and heterogeneous IoV environment. Therefore, our research aims to develop a client selection framework, i.e., FairEquityFL, which ensures participation opportunity for all the clients in such an environment. We ran extensive simulations on FEMNIST dataset and compared our framework's results vis-\`a-vis the baseline FL frameworks. Simulation results depict that our envisaged framework outperforms the baseline FL frameworks to a considerable extent. The salient contributions of the paper-at-hand are as follows:

\begin{itemize}
    \item We propose a FL client selection framework that ensures opportunity for all the clients to participate in the FL model training process in the context of a dynamic and heterogeneous IoV environment. We also introduce a \textit{sampling equalizer module} within the \textit{selector} component that keeps the clients' historical records to optimize the client selection process. 
    \item We propose an outlier detection mechanism based on the model performance in terms of considerable fluctuation in either accuracy or loss minimization. The \textit{client tracker records} tool is responsible for flagging suspicious clients' participation and temporarily suspends them from further participation.
\end{itemize}

The paper is organized as follows. \hyperref[sec:sota]{Section~2} presents state-of-the-art on FL client selection techniques. In \hyperref[sec:sysmodel]{Section~3}, we introduce FairEquityFL framework. \hyperref[sec:expsetup]{Section~4} presents the experimental setup and simulation results. In \hyperref[sec:conclusion]{Section~5}, we conclude the paper and delineate directions for the future research.
 
\section{State-of-the-Art} \label{sec:sota}

Over the past decade, there have been revolutionary changes in the way vehicles are now connected with each other, smart objects, and devices, i.e., road side units, sensors, and actuators \cite{arooj2022big}. A synchronised and real-time connection among the devices, vehicles, and objects within the IoV network can accelerate the coordination, collaboration, and automation in a more efficient and effective way \cite{lai2021oort}. These connected vehicles and devices generate a massive amount of data that needs to be processed to get real-time and fact-driven information safeguarding the privacy of the participating clients \cite{pandya2023federated}. Hence, FL can be utilized to train models without violating individual's privacy since the clients would only share the local gradients \cite{jagarlamudi2024exploring}. Since only a portion of clients can take part in a concurrent round, it is challenging to incorporate fairness and ensure equal opportunity for all clients to participate in FL model training. 

In \cite{zhu2022online}, authors proposed an asynchronous FL online client selection framework considering clients' availability and long-term fairness. This framework converted the latency minimization issue into a multi-armed bandit problem by leveraging the upper confidence bound policy and virtual queuing technique according to the Lyapunov optimization. In \cite{lai2021oort}, the authors proposed Oort, a guided participant selection mechanism in FL to improve time-to-accuracy performance and to achieve sub-linear regret performance. Oort selects the clients that have the data to enhance the global model accuracy and accelerate the training process. Furthermore, it manages the trade-off between the statistical utility and system efficiency. Nevertheless, this paper did not consider fairness while selecting the participating clients. Another study\cite{li2022pyramidfl}, presented PyramidFL to achieve high performance through exploiting the data and system heterogeneity among the selected clients. This framework profiled the clients' utility and prioritized the clients with high statistical as well as system utility. However, PyramidFL did not consider fairness in the FL client selection process.

The authors in \cite{balakrishnan2022diverse} proposed federated averaging with diverse client selection (DivFL)' through the selection of a small portion of clients representing the gradient information. The client's subset was chosen by maximizing the sub-modular facility location function that was defined based on the gradient space. Although DivFL enhanced the FL model performance and achieved faster convergence but did not involve the data from all participants. In \cite{ezzeldin2023fairfed}, authors put forward a new FL framework, FairFed that incorporates group fairness aware FL client selection by utilizing various de-biasing methods. Although this method worked in the scenarios where clients could be easily categorized, however, it did not perform effectively on highly heterogeneous data scenarios. Moreover, in \cite{huang2022stochastic}, stochastic FL client selection method for volatile clients was proposed considering fairness and data weight of the participants. \cite{10219820} presented FairFedCS framework based on Lyapunov optimization to dynamically adjust the probabilities by considering client's reputation, participation time and contribution, however, FairFedCS did not incorporate malicious client detection mechanism.  

In \cite{li2024adafl}, the authors introduced AdaFL to dynamically adjust the clients number utilizing a piece-wise function. In AdaFL, FL training initiated with a small number of clients, which gradually increased with the model's progress. Moreover, this framework evaluated clients' contributions through the performance metrics utilizing weighted average function. Furthermore, in \cite{singhal2024greedy}, the authors presented a biased client selection mechanism, GreedyFed, wherein the highest contributing clients got selected in each round. In \cite{zhou2023hierarchical}, a clustering-based approach was presented utilizing the social context data for FL client selection in Internet of medical things applications. Another study proposed MINDFL framework which prioritized the quality and fairness in FL client selection to deal with challenges related to utility divergence \cite{zhang2023mindfl}. MINDFL assessed the contribution of the local model utilizing the metric called quality-of-model. 

\section{FairEquityFL Framework Design} \label{sec:sysmodel}

In this section, we introduce a new FL framework, FairEquityFL, for the FL client selection process, wherein we intend to provide opportunity for all the clients to participate in the FL model training process in order to realize a fair and equitable client selection in heterogeneous IoV network. To be more specific, FairEquityFL aims to provide each client with precisely tailored opportunities, taking into account their unique circumstances, i.e., frequency of participation, total number of times a client has already participated, and the consecutive number of rounds that the client was not selected for, to achieve equity across clients' participation \cite{mozaffari2022e2fl}. This is realized via a \textit{sampling equalizer module} within the \textit{selector} component. Since the inception of the FL, it has been utilized as a privacy preserving paradigm in the domain of ML \cite{soltani2022survey}. In FL, loss function defines the difference between the prediction and the actual outcomes for the heterogeneous data distribution among the clients. Therefore, an underlying rationale of the FL model is to minimize the loss functions. FL model is trained at clients' level utilizing the respective training samples of the participating clients \cite{zhao2022participant}. Thus, the participating clients' data and corresponding labels contribute towards the model training. Therefore, we define the loss function of the sample $s$ as $f(x_s,y_s,w)$, wherein, $x_s$, $y_s$ and $w$ denotes the vector of the input sample, its corresponding label and existing weight vector, respectively. For the ease of understanding, we utilize $f_s(w)$ to denote $f(x_s,y_s,w)$, therefore, we can express the loss function $F_i (w)$ of the local dataset $D_i$ as below:

\begin{equation}
F_i(w) = \frac{\sum_{s \in D_i} f_s(w)}{|D_i|}
\end{equation}

Likewise, we can express the global loss function $F(w)$ as below: 

\begin{equation}
F(w) = \frac{\sum_{i=1} F_i(w)}{|D|}
\end{equation}\label{Equation 3}

If there are $K$ clients in total, i.e., $C_1, C_2,...C_K$, participating in the model training process, then at the $q^{th}$ round, client $K$ calculates the local updates of its loss function, considering the global model $w$ as depicted in \hyperref[Equation 3]{Equation~3}.  

\begin{equation}
D_w s = \frac{\partial F_s}{\partial w}
\end{equation}

Hereafter, local updates are forwarded to the server and server aggregates the local model for the global model $w$ refinement, where $\alpha_q$ is the learning rate:

\begin{equation}
w(q+1) = w(q) + \alpha_q \frac{1}{N} \sum_{s=1}^{N} \Delta w_s
\end{equation}

\begin{figure}[!t]
    \centering
    \includegraphics[width=\textwidth]{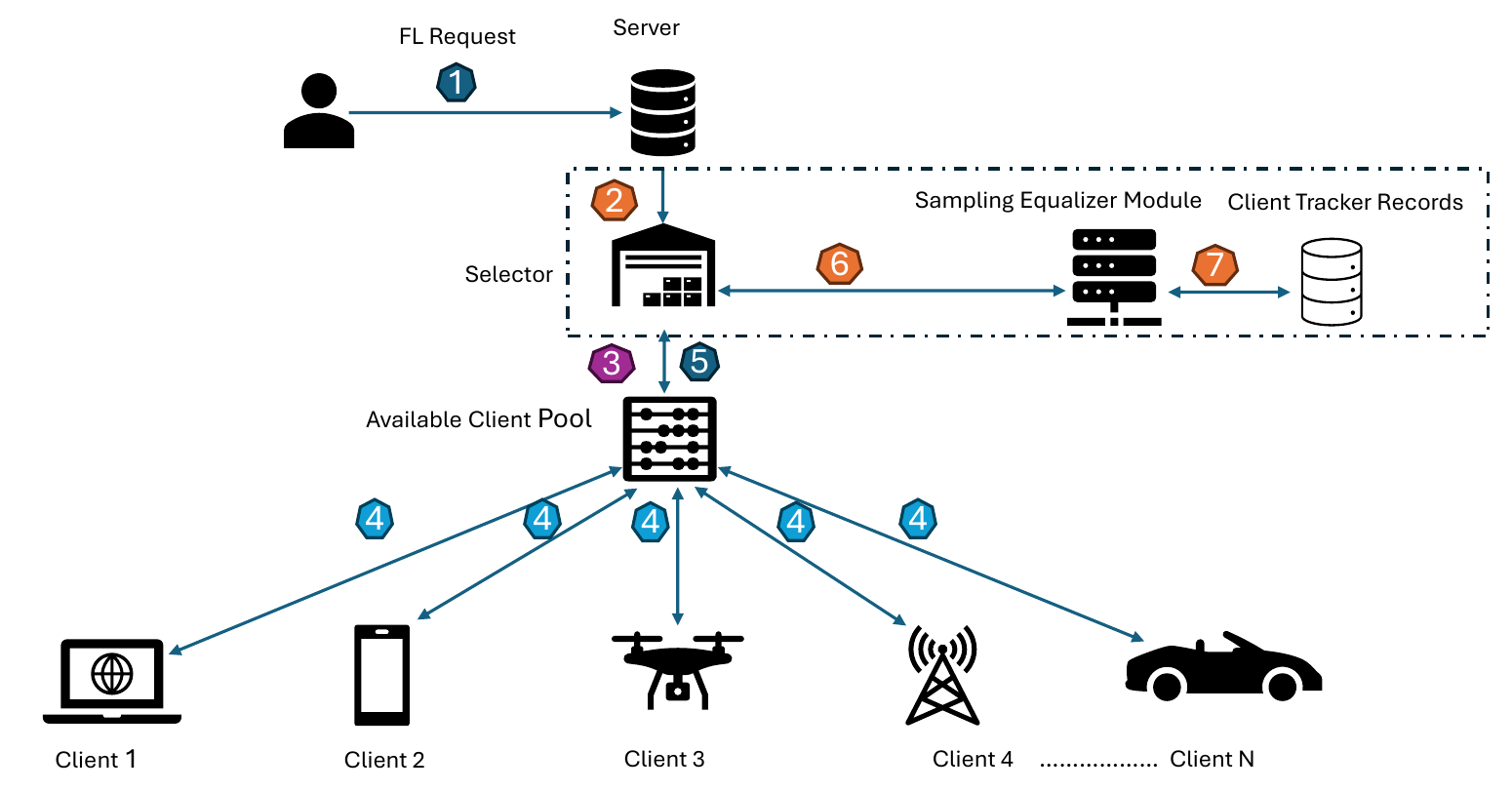}
    \caption{The envisaged FairEquityFL framework.}
    \label{fig:Figure 2}
\end{figure}

Our goal is to minimize the loss function and increase the model accuracy without degrading the time-to-convergence performances and ensure fairness in terms of clients' participation in model training. \hyperref[fig:Figure 2]{Figure~2} presents our proposed FarEquityFL framework. In FairEquityFL, \textit{selector} is responsible for the client selection in each communication round, whereas, the \textit{sampling equalizer module} is responsible for monitoring and controlling the clients' collaboration to ensure equal participation opportunity for all the clients. \textit{Sampling equalizer module} is a part of \textit{selector} component which directly collaborates with the \textit{selector} to (a) guide and harmonizes the client selection process, (b) provide feedback to the \textit{selector}, (c) notify about a client if it is being ignored for a long time or participating too frequently in the consecutive rounds, and (d) flag the client if outlier is detected. Therefore, the \textit{selector} selects the clients with the feedback and recommendation from the \textit{sampling equalizer module}. Within the \textit{sampling equalizer module}, there is a \textit{client tracker records} tool that keeps the record of participating clients in each round. Hence, the \textit{sampling equalizer module} can access clients' historic record of participation, thereby, is responsible to ensure that all clients are participating in the model training in a fair and systematic way to ensure equity. The \textit{sampling equalizer module} is also responsible to ensure that no client is ignored for a long period of time and restrict any client from participating too frequently. 

Therefore, our client selection mechanism is designed in a way that it can control the clients' selection mechanism to ensure equal participation opportunity for all clients during the model training process. Furthermore, our simulations results demonstrates that FairEquityFL improves loss minimization, increases fairness index, model accuracy and time-to-convergence performances. Through the incorporation of the client selection techniques, we have ensured that same client does not participate too many times and no client is ignored for longer time period. Additionally, FL is vulnerable to malicious clients' attacks as the FL server does not have the authority to govern clients' behaviour \cite{li2020learning}. Hence, the malicious clients may attempt to frequently participate in the model training and send inaccurate model updates to the server to negatively affect the local training by initiating model poisoning attack \cite{onsu2023cope}. Therefore, it is critically important that the FL framework should have the defence mechanism against such attacks. Hence, FairEquityFL offers an outlier detection mechanism to flag such suspicious clients and temporarily suspend them from participating. If our proposed model identifies any unusual behaviour of clients, i.e., frequent attempt to participate, considerable fluctuation in accuracy or loss levels in consecutive rounds, it suspends the client from participation and contributes to the outlier detection. Therefore, through identifying unusual performance of clients and by mitigating frequent participation of the clients, our proposed model contributes towards achieving resiliency against malicious clients and enhancing the model security and reliability. 

Our key contribution in the envisaged model lies in incorporating fairness into the client selection process and strengthening the security mechanism via an outlier detection mechanism. Each client can only participate up-to a maximum number of times $ N_{Cmax} $ during the entire FL model training process. This ensures that a client is not selected too many times to (a) avoid biased and inaccurate model result and (b) ensure equity amongst the participation of all clients. $N_{Cmin}$ implies the minimum number of time a client would be selected. The underlying rationale for opting $N_{Cmin}$ is that a client should at least participate once in the FL training process so as to guarantee equity. Here, we define the entire client set as $C$, where we have a total $K$ number of  clients, i.e., $C_i = \{ C_1, C_2, C_3, C_4, C_5, \ldots, C_K \}$. Therefore, in the entire FL training process, the participation of each client is controlled, in such that: 

\begin{equation}
N_{Cmin} <C_i <N_{Cmax} 
\end{equation}

Moreover, a minimum round gap $Gap_{min}$ has been enforced which ensures that no client can participate in two consecutive rounds. For instance, if the client $C_i$ gets selected in the $k^{th}$ round, it would not be re-selected until $(k+ Gap_{min})^{th}$ round. The minimum round gap that a client can be re-selected depends on a number of factors, including but not limited to: total number of clients, total number of rounds, and application type. 

Additionally, FairEquityFL automatically includes a client in a training round, once it is overlooked for a certain number of rounds $Gap_{max}$. This is to ensure that the client has not been overlooked for a long period of time and to guarantee equity in FairEquityFL. However, if the client has already been selected for a maximum permissible number of times, $N_{Cmax}$, it will not be re-selected even though $Gap_{max}$ has reached. If client $C_i$ is selected at the $k^{th}$ round and is overlooked until $(k+ Gap_{max})^{th}$ round, the said client would automatically be re-selected in the succeeding round. Client's historical participation record is stored in the \textit{client tracker records} tool within the \textit{selector} module. Nevertheless, if there are multiple clients reaching $Gap_{max}$ threshold, the model would only select $m$ number of such clients in a single round, whereas, the rest of such clients would be selected in the subsequent rounds in order to make it fair for all the clients and ensure equity in the client selection mechanism. Since the gap limit would be higher for the remaining clients, our model would select the clients in a descending order from such group of clients. Therefore, the higher the gap is, the faster they will be selected to include fairness in the selection mechanism. Furthermore, FairEquityFL eliminates the risk of a client to be ignored during the execution of the FL training. For instance, if a client has not been utilized until a certain interval $C_\lambda$, i.e., $n^{th}$ round, the model would select such a client on priority basis with subject to the participation slots availability $Slots_a$ in a given round. The maximum number of un-utilized clients that can be opted for a particular round is represented as $\lambda_{max} $. It is pertinent to mention that not every $C_\lambda$ clients should be picked in a single round to avoid imbalance in the model.

\begin{algorithm} [H]
\caption{Client selection algorithm}
    \begin{algorithmic}[1]
            \State Parameters: Total Number of Clients - $K$, Number of Selected Clients - $n$, Round Number - $k$, Selected clients at Round $k$ - $C_k$, Not utilized clients - $C_{NU}$, Overlooked clients - $C_G$, Number of times $C_i$ is selected - $T_i$, Number of times $C_i$ is not selected in consecutive rounds - $G_i$, Available slots for both non utilized and overlooked clients - $Slots_a$, Interval at which non-utilized clients are checked - $\lambda$, Maximum number of rounds that a client can be overlooked - $Gap_{max}$, Minimum number of rounds that a client can be reselected - $Gap_{min}$, Maximum number of times that a client can be selected - $N_{Cmax}$, Maximum number of overlooked and non-utilized clients that can be selected for training in a particular round - $m$ , $\lambda_{max} $

  	    \State // Un-utilized clients selected after every given interval
            	\If{ k  \%  $\lambda$ == 0}
       	\State $Max_{NU}$ = min($\lambda_{max} $, $Slots_{a}$)     	 
       	\State  Select $Max_{NU}$ clients ($C_{NUS}$) from the not utilized client pool ($C_{NU}$)
        	\State $C_k \gets  C_{NUS}$
                \State $C_{NU} \gets \{c \in C_{NU} \,|\, c \notin C_{NUS}\}$
                \State $Slots_a \gets  Slots_a - max_u$
         
      	\EndIf
    	\State // Clients selected after reaching a maximum gap
     	\State $ counter \gets  0$
     	\For{$C_i \in C_G$} 
            \State $max_{o}$ = min($m $, $Slots_{a}$)  
       	\If{$counter <= max_{o}$}
            	\If{$G_i>=Gap_{max}$ and $T_{i}<N_{Cmax}$}
                   	\State $counter \gets counter + 1$
                   	\State $C_k \gets  C_k \cup  C_i$
            	\EndIf
        	\EndIf
    	\EndFor  
     
    \State // Filtering the clients that cannot be selected due to the not reaching of $Gap_{min}$

    \State $C_{available_1} \gets \{c \in C_K \,|\, c \notin C_k\}$

    \For{$C_i \in C_{available_1}$}
        \If{$G_i$ >= $Gap_{min}$}
            \State $C_{available_2} \gets  C_{available_2} \cup  C_i$
        \EndIf
    \EndFor
    
    \State // Remaining number of clients are selected randomly

            \State $n_r \gets  n - Count(C_k)$
            \State $C_{rand} \gets  Rand(C_{available_2}, n_r)$
            \State $C_k \gets  C_{rand} \cup  C_k$
        
     	\State // Recalculation of $T_i$ and $G_i$ of clients
                \For{$C_i \in C_K$}
       	        \If{$C_i$ in $C_k$}
                        \State $T_i \gets T_i + 1$  
        	        \State $G_i \gets 0$  
                    \Else
                        \State $G_i \gets G_i + 1$  
       	        \EndIf
    	    \EndFor
    \end{algorithmic}
\end{algorithm}

Finally, in scenarios, wherein either model accuracy decreases $Acc_{th}$\% or loss minimization decreases $Loss_{th}$\% considerably, FairEquityFL identifies and suspends such common reaching threshold level clients for a certain number of rounds $X_n$. For instance, assuming client $C_1$ gets selected in $X_n$ number of rounds, wherein, the accuracy level decrease reach the threshold $Acc_{th}$\%, the model would suspend $C_1$ for next $ST_n$ rounds from further participation in model training. Security mechanism in FairEquityFL aids to detect suspicious clients to strengthen the reliability of FL model.

\section{Experimental Setup and Simulation Results} \label{sec:expsetup}

To simulate and evaluate the performance of FairEquityFL, we have utilized one of the most commonly used datasets in FL, FEMNIST, so that we can compare the simulation results with the state-of-the-arts models, i.e., FedAvg, FedProx, CSFedAvg and Newt. FEMNIST dataset is the federated version of MNIST dataset. Each of the images in FEMNIST is 28 x 28 pixels in size and the format is gray-scale. For creating data heterogeneity, we have sorted the dataset to create subset of data for each class and divided each label's data into smaller shards where each shard has mix images from different classes. We have randomly allocated the shards among the clients in FEMNIST dataset. 

\subsection{Performance in Terms of Fairness}

We have incorporated Jain's Fairness Index (JFI) to measure the fairness level in FL client selection so that the clients with heterogeneous data can be fairly selected for the model training \cite{shi2023fairness}. Here, $S_i$ indicates the total number of times client $i$ participated in the model training and $Q_i$ represents the quality of the data in client $i$. For the evaluation of client's data quality, We have utilized the method presented in \cite{deng2021auction}. So the formulation is as follows:

\begin{equation}
   \text{JFI} = \frac{\left( \sum_{i=1}^{n} \frac{S_i}{Q_i} \right)^2}{n \sum_{i=1}^{n} \left( \frac{S_i}{Q_i} \right)^2} 
\end{equation}

Here, the quality of data for client $i$ is determined by two factors. Firstly, the number of image classes ($n_{class}$) that the client has and secondly, the proportion of the client's data that is noisy ($P_{noisy}$). Therefore, the formula can be expressed as follows:

\begin{equation}
    Q_i = n_{\text{class}_i} \times (1 - p_{\text{noisy}_i})
\end{equation}

We have normalized the $Q_i$ from [0.5, 1] to [0, 1] to ensure that data quality scores are comparable across noisy and heterogeneous data settings for fair client selection process. JFI ranges from 0 (indicates most unfair) to 1 (indicates most fair). We have run the simulation multiple times in order to calculate the mean JFI. \hyperref[tab:jfi_values]{Table~1} depicts that FairEquityFL demonstrates higher JFI compared to the baseline models as most of these models have not considered equity in terms of clients participation. However, achieving the fairness level 1 is almost impossible as the model would be impacted by un-availability of the clients and with the restriction of the suspicious clients' participation. 

\begin{table}[!t]
    \centering
    \caption{Jain's Fairness Index (JFI) for the baseline FL frameworks vis-\`a-vis our envisaged FairEquityFL framework.}
    \begin{tabular}{>{\centering\arraybackslash}p{2.2cm}|>{\centering\arraybackslash}p{1.8cm}|>{\centering\arraybackslash}p{1.8cm}|>{\centering\arraybackslash}p{1.8cm}|>{\centering\arraybackslash}p{1.8cm}|>{\centering\arraybackslash}p{2.2cm}}
        \hline\hline
        \textbf{Algorithm} & \textbf{FedAvg} & \textbf{FedProx} & \textbf{CsFedAvg} & \textbf{Newt} & \textbf{FairEquityFL} \\
        \hline
        \textbf{JFI} & 0.539 & 0.398 & 0.426 & 0.510 & 0.816 \\
        \hline\hline
    \end{tabular}
    \label{tab:jfi_values}
\end{table}

\subsection{Performance in Terms of Accuracy Level}

\begin{figure}[H]
    \centering
    \includegraphics[width=0.7\textwidth]{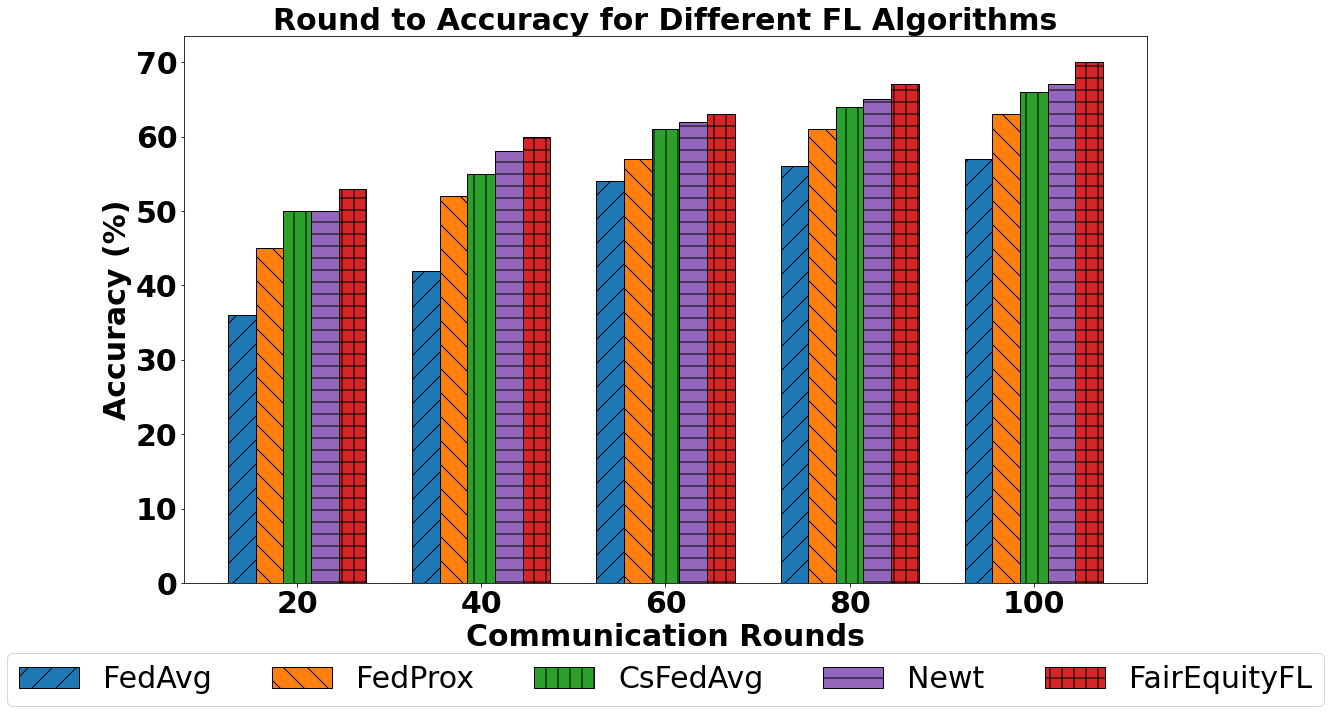}
    \caption{Round-to-accuracy for the baseline FL frameworks vis-\`a-vis our envisaged FairEquityFL framework.}
    \label{fig:Round to Accuracy}
\end{figure}

 We ran FairEquityFL simulation for 100 rounds multiple times to realize performance consistency and stability. \hyperref[fig:Round to Accuracy]{Figure~5} depicts the accuracy level comparison for FairEquityFL with baseline frameworks. For the ease of understanding, we presented the accuracy level comparison at every 20\textsuperscript{th} rounds and the result depicts, FairEquityFL outperforms the baseline frameworks in terms of round-to-accuracy at every stage. 

 Among the baseline frameworks, maximum accuracy of FedAvg is the lowest, i.e., 57\%. To achieve this 57\% accuracy FedAvg, FedProx, CSFedAvg, Newt and FairEquityFL has taken 100, 66, 50, 43 and 34 rounds respectively utilizing the same dataset FEMNIST. For FairEquityFL, maximum accuracy level is 70.86\%, whereas, for Newt, CSFedAvg, FedProx and FedAvg the maximum accuracy is 67.28\%, 66.26\%, 63.02\%, and 57.66\% respectively.   

 \subsection{Performance in Terms of Time-to-Convergence}

\begin{figure}[H]
    \centering
    \includegraphics[width=0.7\textwidth]{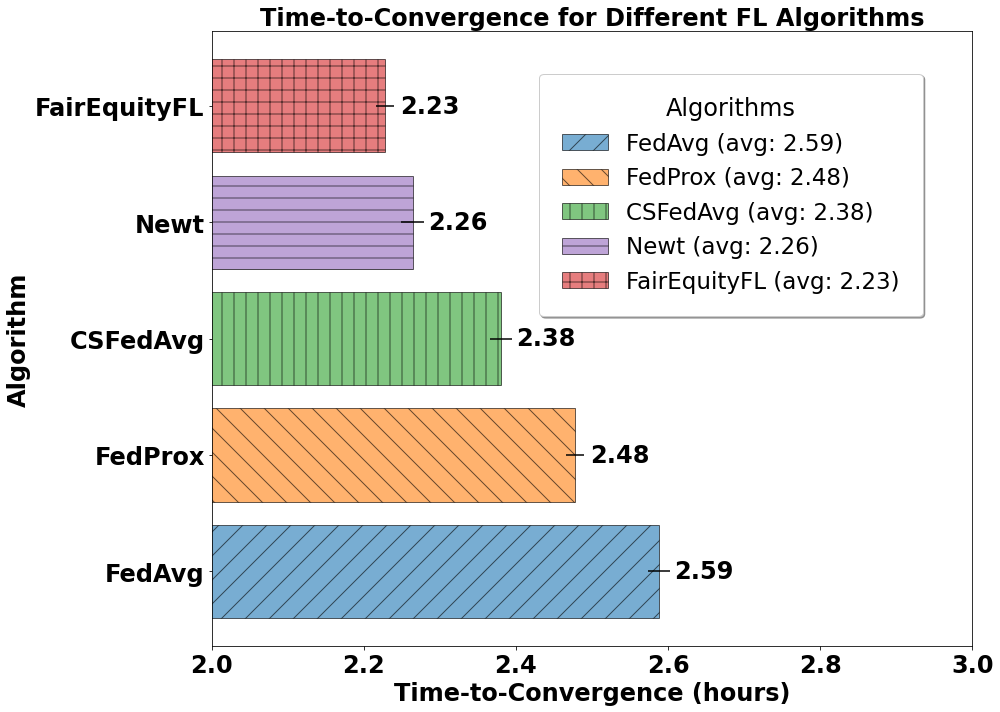}
    \caption{Time-to-convergence comparison for the baseline FL frameworks vis-\`a-vis our envisaged FairEquityFL framework.}
    \label{Time-to-convergence comparison}
\end{figure}

We evaluated FairEquityFL performance to reach the optimum accuracy not only in terms of the number of communication rounds but also considered the time-to-convergence to measure the systems performance. In IoV network, there are a number of time-sensitive applications, i.e., autonomous driving, collision avoidance, and traffic monitoring systems. Therefore, time-to-convergence performance is crucial for such time critical applications. The earlier the model would achieve convergence and reach the optimum accuracy the better the model would be able to serve the time-sensitive applications. Following the same procedure as stated in \cite{zhang2021client}, we recorded the time and round number, FairEquityFL took to achieve the optimum accuracy and convergence. We measured the system performance from two aspects, i.e., the accuracy level increased in a single round and the time duration of each round. These two aspects, together guided us to measure the time-to-accuracy performance. We have denoted the time-to-accuracy and round-to-accuracy parameters as $TA_{Accuracy}$ and $RA_{Accuracy}$ respectively, wherein A is our optimum accuracy, that is 70.86\%. Likewise, we have measured and recorded the time and number of rounds it took to achieve the maximum accuracy for the state-of-the-arts models. Figure \ref{Time-to-convergence comparison} states the time-to-convergence performance to achieve the optimum accuracy for the baseline algorithms. As we can see, FedAvg, FedProx, CSFedAvg and Newt achieve the convergence in 2.59, 2.48, 2.38, and 2.26 hours, whereas FairEquityFL framework achieves the convergence in 2.23 hours. This results represents FairEquityFL outperforms the baseline frameworks in terms of time-to-convergence rate with higher accuracy.

 \subsection{Performance in Terms of Loss Minimization}

\begin{figure}[H]
    \centering
    \includegraphics[width=0.7\textwidth]{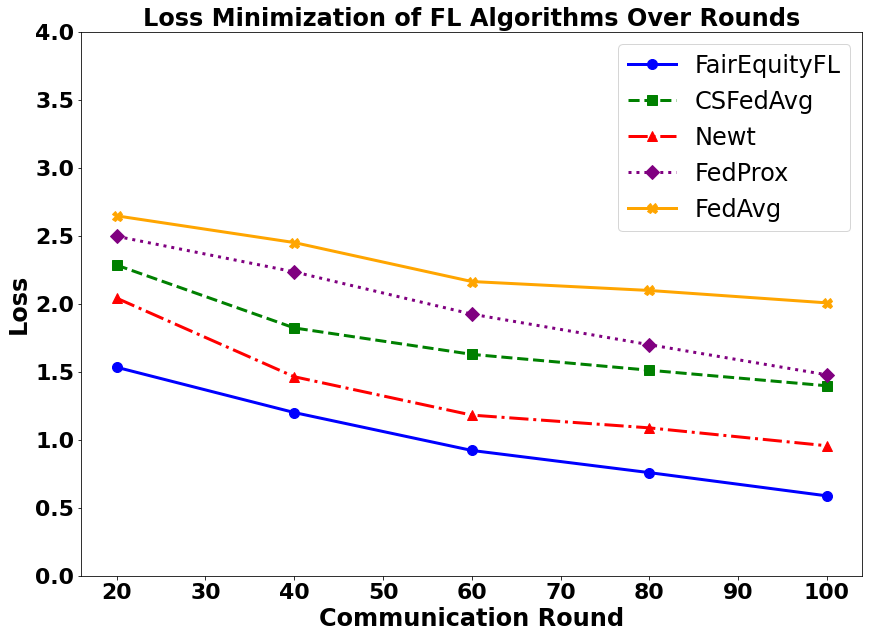}
    \caption{Communication rounds vs. loss minimization for the baseline FL frameworks vis-\`a-vis our envisaged FairEquityFL framework.}
    \label{Loss Comparison}
\end{figure}

Loss minimization is crucial in FL, as it is directly related to the efficiency and quality of the aggregated global model. Moreover, it enhances the predictive accuracy of FL model and assures that each client gets benefited from the aggregated global model. Hence, prioritizing loss minimization in heterogeneous IoV network is pivotal. Simulation results depicts that FairEquityFL achieves higher loss minimization compared to baseline frameworks. FairEquityFL temporarily suspends the clients which demonstrates higher decrease in loss minimization, thereby, contributes to loss minimization by restricting the participation of such clients. Figure \ref{Loss Comparison} presents a comparison of FairEquityFL loss minimization and demonstrates that FairEquityFL achieves higher loss minimization factors compared to the baseline frameworks.

\subsection{Performance in Terms of Outlier Detection}

\begin{figure}[H]
    \centering
    \includegraphics[width=0.7\textwidth]{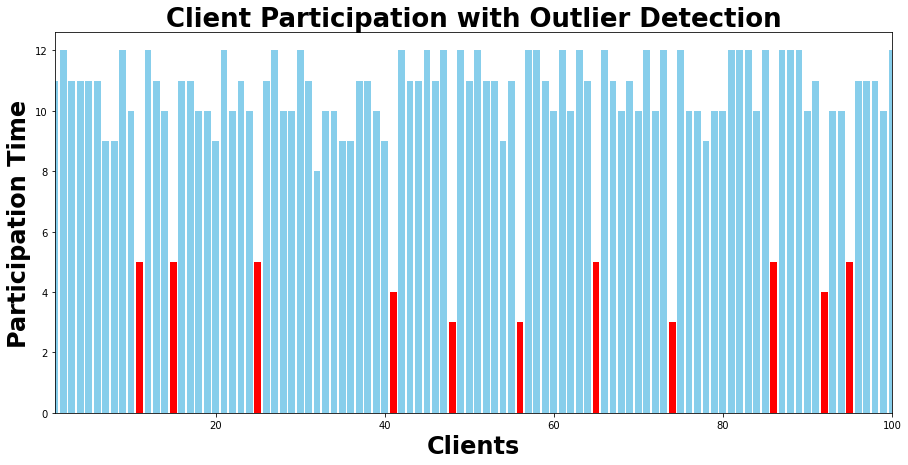}
    \caption{Outlier detection for the envisaged FairEquityFL framework.}
    \label{fig:outlier}
\end{figure}

For evaluating FairEquityFL performance in terms of outlier detection, we ran the simulation and observed if FairEquityFL can successfully identify the clients that reaches the threshold of decrease in accuracy $Acc_{th}$\% or loss minimization $Loss_{th}$\%. We had tested multiple times with different threshold values to observe the performance consistency. The simulation results depicts, FairEquityFL successfully identified the clients reaching the threshold and temporarily suspend those suspicious clients from further participation in model training. \hyperref[fig:outlier]{Figure~3} represents clients which had been identified as suspicious with red color. Though the temporary suspension of suspicious clients would affect fairness in terms of equal participation opportunity, however, it would enhance the systems performance, security and reliability. Therefore, we considered it as trade-offs between the system performance and fairness.

\section{Conclusion and Future Directions} \label{sec:conclusion}

With the rapid proliferation of the smart objects, sensors and devices, IoV network has been greatly flourished in the recent years. FL can significantly contribute towards deriving real-time, fact-based information and safeguarding the privacy of participating clients. Therefore, FL has been incorporated extensively in IoV network. Since FL model is developed based on the gradients shared by the participating clients, it is crucial to incorporate fairness into FL client selection process. In this paper, we present FairEquityFL, a FL client selection mechanism that ensures opportunity for all clients to participate in the model training. Compared to the existing frameworks, FairEquityFL demonstrates high level fairness incorporation, accuracy enhancement with loss minimization, improvement in time-to-convergence and round-to-convergence parameters. Moreover, we introduce a \textit{sampling equalizer module} and \textit{client tracker records} tool that can monitor and control clients' participation in a fair and systematic way. Furthermore, our proposed approach includes an outlier detection mechanism that can flag the suspicious clients based on considerable fluctuation in accuracy level decrease or loss minimization decrease. In this work, we have incorporated the logic to temporarily suspend the suspicious clients from participating in model training. However, further research is required to confirm the level of maliciousness of the suspicious clients. We will investigate about this further to build an intelligent and robust security mechanism to select trustworthy clients for FL model training in our future work. 
\clearpage
\bibliographystyle{ieeetr}
\clearpage
\bibliography{Techbib}

\end{document}